\documentclass[10pt,twocolumn,letterpaper]{article}

\usepackage[pagenumbers]{cvpr} 

\definecolor{cvprblue}{rgb}{0.21,0.49,0.74}
\usepackage[pagebackref,breaklinks,colorlinks,allcolors=cvprblue]{hyperref}
\usepackage[capitalize]{cleveref}
\usepackage{booktabs}
\usepackage{amssymb}
\usepackage{multirow}
\usepackage{wrapfig}
\usepackage{academicons} 
\definecolor{LightBlue}{rgb}{0.9,0.94,1}

\def\paperID{5951} 
\def\confName{CVPR}
\def\confYear{2025}

\title{CatV$^{2}$TON: Taming Diffusion Transformers for Vision-Based Virtual Try-On with Temporal Concatenation}

\author{
    \textbf{Zheng Chong}\textsuperscript{\rm 1,4*},
    \textbf{Wenqing Zhang}\textsuperscript{\rm 2*}, 
    \textbf{Shiyue Zhang}\textsuperscript{\rm 1}, 
    \textbf{Jun Zheng}\textsuperscript{\rm 1}, 
    \textbf{Xiao Dong}\textsuperscript{\rm 1},\\
    \textbf{Haoxiang Li} \textsuperscript{\rm 3},
    \textbf{Yiling Wu}\textsuperscript{\rm 4}, 
    \textbf{Dongmei Jiang}\textsuperscript{\rm 4}, 
    \textbf{Xiaodan Liang}\textsuperscript{\rm 1,4$\dagger$}
    \vspace{2mm}
    \\
    Sun Yat-Sen University\textsuperscript{\rm 1}, National University of Singapore\textsuperscript{\rm 2}, \\
    Pixocial Technology\textsuperscript{\rm 3}, Pengcheng Laboratory\textsuperscript{\rm 4}   
    \vspace{1mm}\\
    \small\texttt{* Equal Contribution,    $\dagger$ Corresponding Author} \vspace{1mm}\\
    \noindent\makebox[\linewidth][c]{\normalsize\href{https://github.com/Zheng-Chong/CatV2TON}{\texttt{https://github.com/Zheng-Chong/CatV$^{2}$TON}}}
}

\begin{document}

\twocolumn[{
    \renewcommand\twocolumn[1][]{#1}%
    \maketitle
    \centering
    \captionsetup{type=figure}
    \includegraphics[width=\textwidth]{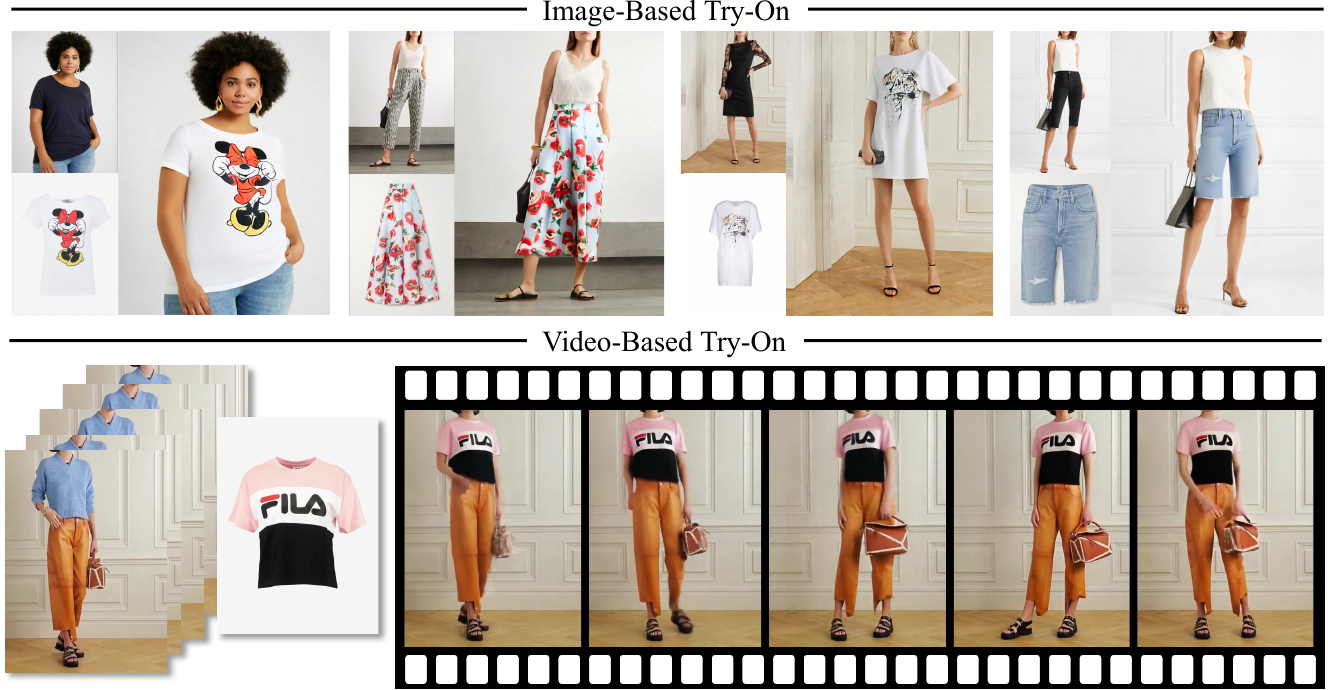}
    \captionof{figure}{Examples of CatV$^2$TON's unified virtual try-on capabilities, demonstrating high-quality garment consistency across both image-based and video-based try-on tasks, including dynamic long-video scenarios.
    }
    \label{fig:teaser} 
    \vspace{4mm}
}]
\begin{abstract}

Virtual try-on (VTON) technology has gained attention due to its potential to transform online retail by enabling realistic clothing visualization of images and videos. However, most existing methods struggle to achieve high-quality results across image and video try-on tasks, especially in long video scenarios.
In this work, we introduce CatV$^2$TON, a simple and effective vision-based virtual try-on (V$^2$TON) method that supports both image and video try-on tasks with a single diffusion transformer model. By temporally concatenating garment and person inputs and training on a mix of image and video datasets, CatV$^2$TON achieves robust try-on performance across static and dynamic settings. For efficient long-video generation, we propose an overlapping clip-based inference strategy that uses sequential frame guidance and Adaptive Clip Normalization (AdaCN) to maintain temporal consistency with reduced resource demands. We also present ViViD-S, a refined video try-on dataset, achieved by filtering back-facing frames and applying 3D mask smoothing for enhanced temporal consistency. Comprehensive experiments demonstrate that CatV$^2$TON outperforms existing methods in both image and video try-on tasks, offering a versatile and reliable solution for realistic virtual try-ons across diverse scenarios.
\end{abstract}  
\section{Introduction}
\label{sec:intro}

The rapid evolution of image and video synthesis techniques has driven significant advancements in downstream tasks, including vision-based virtual try-on, which can be categorized into image-based and video-based approaches. Image-based virtual try-on methods \cite{xu2024ootdiffusion, wang2024stablegarment, sun2024outfitanyone, chong2024catvton, xie2023gpvton, choi2024idmvton, zhang2024mmtryon, li2023warpdiffusion, xie2022pastagan++} have been extensively explored, achieving high levels of garment realism and detail on static images. Meanwhile, video-based virtual try-on \cite{fang2024vivid, jiang2022clothformer, he2024wildvidfit, fwgan-vvt, xu2024tunnel}, bolstered by recent advancements in video generation, has garnered increasing research interest for dynamic applications.

However, current methods for image-based and video-based try-ons often rely on separate model designs and frameworks. For instance, warping-based methods \cite{li2023warpdiffusion, gou2023dcivton, xie2023gpvton, xie2022pastagan++} are tailored specifically for image try-on, adjusting garment shapes to match each pose but unable to maintain temporal consistency across frames. Methods that utilize specialized networks like ReferenceNet or GarmentNet \cite{zhang2024mmtryon, xu2024ootdiffusion, wang2024stablegarment, shen2024imagdressing, sun2024outfitanyone} achieve highly realistic try-on effects, but the added encoding networks increase computational overhead. While video-based try-on methods \cite{he2024wildvidfit, fang2024vivid, xu2024tunnel} have made progress and can perform image try-on tasks, their performance still lags behind models designed exclusively for image try-on. Based on these observations, we introduce CatV$^2$TON, a streamlined vision-based virtual try-on diffusion transformer framework. By temporally concatenating garment and person inputs and training on a combination of image and video datasets, CatV$^2$TON addresses both static and dynamic try-on scenarios within a single, cohesive model. With only 20\% of the backbone parameters allocated to trainable components and no additional modules, it offers a flexible and efficient framework for diverse try-on applications.

Besides, a critical challenge in video generation is producing long, temporally consistent sequences, which is often resource-intensive and susceptible to quality degradation over time. To generate high-quality long videos, many methods \cite{xu2024easyanimate, chai2023stablevideo} typically involve pre-training on short videos followed by fine-tuning on a limited set of long video data. However, the high hardware requirements and extended inference times needed to produce high-definition long videos significantly restrict their practical application.
To address this, we propose an overlapping clip-based inference strategy that leverages preceding frames as temporal guidance and integrates Adaptive Clip Normalization (AdaCN) to ensure consistency while reducing resource costs. Specifically, during training, we use only short video segments with a probability of exposing preceding frames to the model, enabling frame-guided generation. At inference, long video sequences are generated in segments, with the final frames of the previous segment serving as guidance for the next. However, this sequential approach may introduce flickering and color mismatches between segments. To counteract these issues, Adaptive Clip Normalization (AdaCN) is introduced to rectify segments based on guiding frames, thereby maintaining coherence in segmented long video generation.

Additionally, we identified several issues with current video try-on datasets. On the one hand, video try-on datasets include human videos with turning or rotating actions but usually contain only garment images in front view. Due to the lack of back-view garment information, generating realistic try-on results when the person is in a rear-facing pose becomes unfeasible. This discrepancy is particularly noticeable for garments with logos, text, or intricate designs, and Using these back-facing frames as ground truth in training will slow model convergence and reduce garment consistency. To address this, we trained a specialized recognition model to detect the person’s orientation in each frame, filtering out back-facing frames to create a cleaner dataset.
On the other hand, while large quantities of continuous person videos are available as video try-on training data, the indispensable clothing-agnostic masks still rely on frame-by-frame processing using image-based parsing models, which introduces temporal discontinuities. To reduce flickering and edge inconsistency caused by this frame-by-frame processing, we propose a 3D Mask Smoothing operation. This involves applying spatial and temporal average pooling to the clothing-agnostic masks followed by re-binarization, effectively reducing flicker and leakage of edge information across frames.

In summary, the contributions of this work include:
\begin{itemize}
    \item We introduce CatV$^2$TON, a simple yet efficient vision-based virtual try-on diffusion transformer that seamlessly handles both static and dynamic try-on scenarios. It is trained on a mixed image-video dataset with more than 80\% parameters of the backbone frozen by temporally concatenating garment and person inputs.
    \item We propose an overlapping clip-based inference strategy for long try-on video generation, utilizing preceding frames as guidance and applying Adaptive Clip Normalization (AdaCN) to reduce resource demands while maintaining temporal consistency.
    \item We present ViViD-S, a refined video try-on dataset that has undergone noise data reduction and quality enhancement through specially trained recognition models and 3D mask smoothing, providing high-quality samples suitable for video try-on tasks.
    \item Extensive qualitative and quantitative evaluations on both image and video try-on datasets demonstrate that our approach outperforms existing baseline methods in both quantitative and qualitative analysis, as well as in in-the-wild scenarios.
\end{itemize}

\section{Related Work}
\label{sec:related_works}
\subsection{Video Synthesis and Generation}

Significant advancements have been made in video synthesis and generation, especially in generating coherent and high-quality video sequences from text descriptions. Models like StableVideo \cite{chai2023stablevideo} and Dreamix \cite{molad2023dreamix} leverage diffusion models to capture both content and style with temporal consistency, while ImagenVideo \cite{ho2022imagen} and Make-A-Video \cite{singer2022make} focus on high-resolution outputs with fine details across frames. Methods such as FateZero \cite{qi2023fatezero} and PVDM \cite{yu2023video} emphasize inter-frame coherence, crucial for natural-looking animations.
For portrait-relevant editing, maintaining identity and expression consistency across frames remains a key challenge. Approaches like \cite{esser2023structure} balance structural and content control to preserve facial identity, while MagicVideo \cite{zhou2022magicvideo} utilizes latent diffusion for smooth and temporally stable animations. Methods like Tune-A-Video \cite{wu2023tune} and Follow-Your-Pose \cite{ma2024follow} bring innovations in one-shot tuning and pose-guided generation, respectively, for realistic human animations in videos. Additionally, image animation methods like EasyAnimate \cite{xu2024easyanimate}, MagicAnimate \cite{xu2024magicanimate}, and AnimateAnyone \cite{hu2024animate} use transformer-based and reference-guided techniques to enhance temporal coherence and identity preservation for character-specific animations. However, despite these remarkable achievements, high-quality video synthesis still faces challenges in maintaining fine-grained detail consistency, particularly in subject-driven video generation.

\subsection{Vision-based Virtual Try-On}
Vision-based virtual try-on includes both image-based and video-based approaches. Image-based try-on generates realistic garment fittings on a target person's photo from a garment image.
Methods such as OOTDiffusion \cite{xu2024ootdiffusion}, IDM-VTON \cite{choi2024idmvton}, StableGarment \cite{wang2024stablegarment}, and OutfitAnyone \cite{sun2024outfitanyone}, utilize diffusion models and dual-stream networks to achieve high-fidelity garment rendering, addressing challenges like pose variation and complex backgrounds. Methods like GP-VTON \cite{xie2023gpvton}, DCI-VTON \cite{gou2023dcivton}, WarpDiffusion \cite{li2023warpdiffusion}, and GarDiff \cite{wan2024GarDiff} use pre-warped garments as guidance to enhance realism and detail. Approaches like MMTryon \cite{zhang2024mmtryon}, IMAGDressing-v1 \cite{shen2024imagdressing}, and Wear-Any-Way \cite{chen2024wear} allow for user-controlled try-ons with multimodal conditioning. Lightweight solutions, such as CatVTON \cite{chong2024catvton}, provide efficient methods by spatially concatenating garment and person images.
Video-based try-on techniques aim to address temporal consistency and garment deformation across frames. Flow-based methods such as FW-GAN \cite{dong2019fw} and ClothFormer \cite{jiang2022clothformer} employ warping modules to manage garment alignment and occlusions over various poses and backgrounds. Diffusion-based models like ViViD \cite{fang2024vivid}, WildVidFit \cite{he2024wildvidfit}, and VITON-DiT \cite{zheng2024viton} integrate garment encoding and pose tracking to ensure consistent fitting across diverse body movements. Meanwhile, approaches like Tunnel Try-On \cite{xu2024tunnel} enhance motion stability using Kalman filtering for commercially viable, smooth, and detailed garment displays. 
Despite these advancements, current methods are typically limited to single-domain applications, focusing solely on either image or video-based try-ons. A unified vision-based try-on model, capable of seamless virtual garment fitting across both images and videos, remains an area for further exploration.

\section{Method}
\label{sec:methods}

\begin{figure*}
  \centering
  \includegraphics[width=\textwidth]{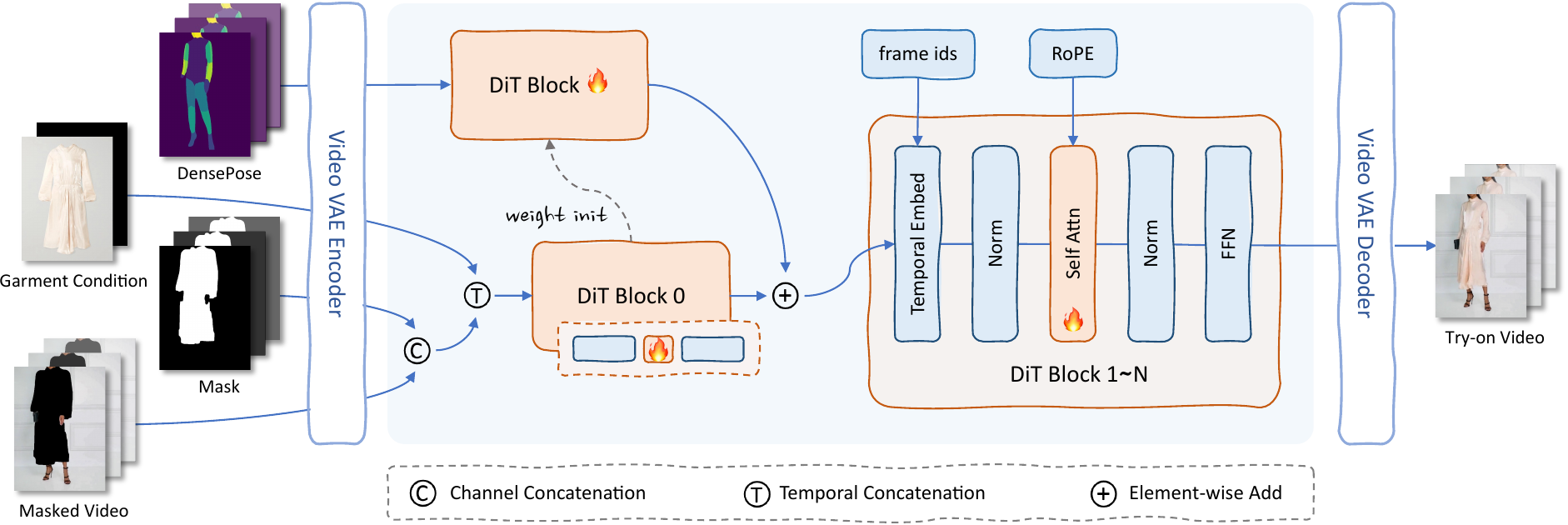}
  \caption{Overview of the CatV$^2$TON architecture. CatV$^2$TON uses DiT \cite{Peebles2022DiT} as the backbone, with the first DiT block duplicated as the Pose Encoder. The person and garment conditions are concatenated temporally as try-on conditions. The entire trainable portion consists only of the self-attention layers and Pose Encoder, accounting for less than 1/5 of the total parameters.}
  \label{fig:framework}
\end{figure*}

Our approach aims to develop a streamlined, efficient vision-based virtual try-on network that addresses high resource demands and continuity challenges in generating extended try-on videos. To this end, we propose a refined diffusion transformer model based on a pre-trained video generation model \cite{xu2024easyanimate}, achieving task adaptation with less than 20\% of the parameters required for full training (see \Cref{subsec:dit}). Furthermore, we introduce a novel Overlapping Clip-Based Inference strategy along with Adaptive Clip Normalization (AdaCN), facilitating the segmented generation of long-form videos (see \Cref{subsec:infer}).
\subsection{Vision-based Try-On Diffusion Transformer}
\label{subsec:dit}

\subsubsection{Input Conditions}
As shown in \Cref{fig:framework}, CatV$^2$TON takes as input images or videos of persons, clothing-agnostic masks, pose representations, and target garment images. These inputs are encoded by the video VAE encoder and projected into the latent space. The main backbone, DiT \cite{Peebles2022DiT}, generates the try-on result through multiple denoising steps, which is then decoded into a video by the video VAE decoder.

\noindent \textbf{Person-Related Conditions.} We apply the mask to occlude the input person video or image, resulting in a masked person representation. This masked person is concatenated with the mask along the channel dimension as conditioning information.

\noindent \textbf{Garment-Related Conditions.} We use an all-zero mask concatenated along the channel dimension to ensure alignment with the person input. The garment conditions are then concatenated with the person conditions along the temporal dimension. Pose guidance is crucial for maintaining motion continuity in dynamic video generation.

\noindent \textbf{Pose Conditions.} We use DensePose \cite{güler2018densepose} as the pose representation, which provides more detailed information compared to skeleton-based methods like OpenPose \cite{openpose} and MMPose \cite{mmpose2020}.

\subsubsection{Network Structure}

State-of-the-art video generation models \cite{xu2024easyanimate, openpose} commonly use Diffusion Transformers (DiTs) \cite{Peebles2022DiT} as the backbone. To leverage pre-trained video generation models and speed up training, we adopt DiT as our backbone and initialize our weights from EasyAnimateV4 \cite{xu2024easyanimate}, removing the cross-attention layers. This modification is made because our task does not require CLIP \cite{radford2021learningtransferablevisualmodels} or text conditions. Our network consists of $N$ stacked DiT blocks, each with temporal and positional embeddings (using RoPE \cite{su2023rope}), and includes only self-attention, feedforward, and normalization layers.
For the Pose Encoder, inspired by ControlNeXt \cite{peng2024controlnextpowerfulefficientcontrol}, we encode DensePose \cite{güler2018densepose} sequences by duplicating the first DiT block and unlocking training, then injecting it into the network using element-wise addition after the first backbone block. This approach is lightweight compared to ControlNeXt \cite{peng2024controlnextpowerfulefficientcontrol}, as our pose encoder is integrated into the backbone, eliminating the need for feature normalization before element-wise addition.

\subsubsection{Training Strategy}
\label{sec:train}
Due to the use of pre-trained weights, and inspired by \cite{chong2024catvton}, we unlock only the self-attention layers that involve interactions. The Pose Encoder is fully unlocked, with only 89.90 M trainable parameters, so the entire trainable portion accounts for less than 20\% of the backbone network's total parameters.
During training, we apply a 10\% chance of dropping the garment condition, which enhances generation quality during inference by enabling classifier-free guidance \cite{ho2022classifierfreediffusionguidance}. To facilitate the generation of long videos in segments, we do not mask the first $k$ frames of the person video during training, setting their masks to all-zero. This allows the model to learn the ability to generate subsequent frames based on the preceding frames. In our experiments, this probability is set to 20\%.

\subsection{Overlapping Clip-Based Inference}
\label{subsec:infer}

To address the high computational resource requirements of generating long video try-ons, we propose the Overlapping Clip-Based Inference strategy. Specifically, during training, we actively expose earlier frames as prompt frames to enable the model to learn the ability to continue generation (as described in \Cref{sec:train}). This allows the model to use the results from previous inference steps as prompts to continue generating during inference.

\begin{figure*}
  \centering
  \begin{subfigure}{0.74\linewidth}
    \includegraphics[width=\linewidth]{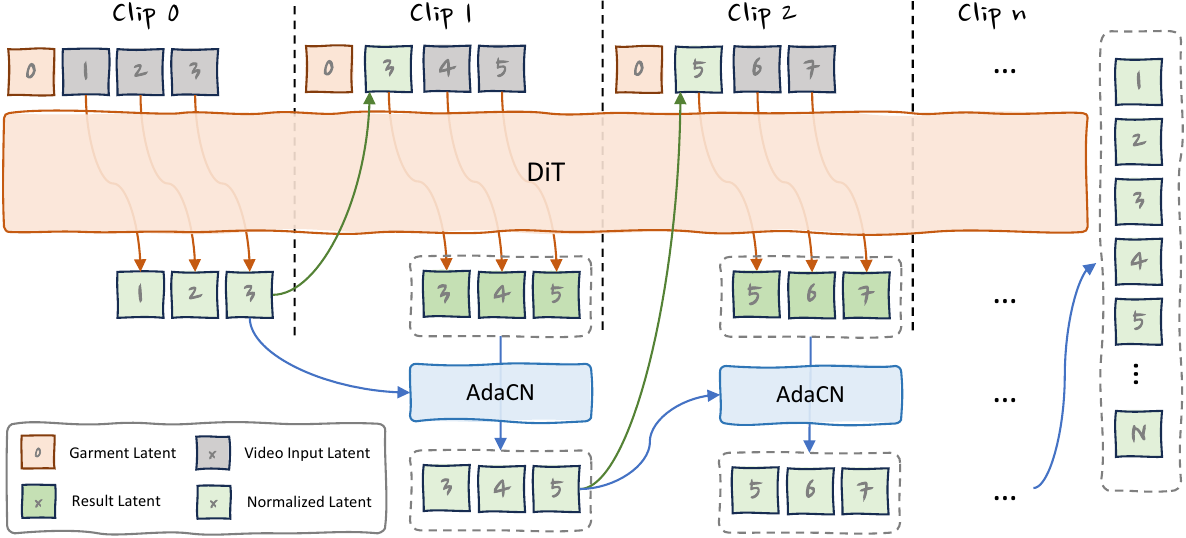}
    \caption{Overlapping Clip-based Inference}
    \label{fig:short-a}
  \end{subfigure}
  \hspace{0.01\linewidth} 
  \vrule width 0.5pt
  \hfill
  \begin{subfigure}{0.22\linewidth}
    \includegraphics[width=\linewidth]{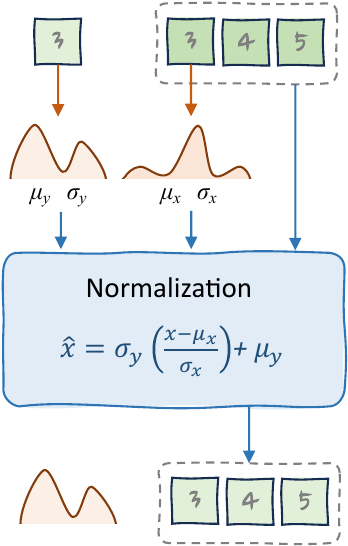}
    \caption{AdaCN}
    \label{fig:short-b}
  \end{subfigure}
  \caption{Illustration of the Overlapping Clip-Based Inference strategy. (a) A long video is divided into $n$ overlapping clips, with each clip consisting of repeated frames. The last $k$ frames of each clip are used as prompt frames for generating the next clip. (b) Adaptive Clip Normalization (AdaCN) is applied to normalize the entire clip based on the mean and standard deviation of the prompt frame features and the denoised prompt frames, ensuring smooth continuity across clips in the generated video.}
  \label{fig:short}
\end{figure*}

In detail, as shown in \Cref{fig:short-a}, for a long video that requires try-on generation, we first divide it into \( n \) clips, each containing repeated frames. For each clip, all frames are masked during inference, and after generating the results, the last \( k \) frames of the clip are used as prompt frames for generating the next clip. This process repeats for the entire video.

However, this inference method presents a challenge: when generating sequentially, the prompt frames, after denoising, may undergo feature shifts, leading to color discrepancies and motion misalignment. This results in relative discontinuities between clips in the final generated video.

To address this issue, inspired by AdaIN \cite{huang2017arbitrarystyletransferrealtime}, we extend it to the level of video clips and propose Adaptive Clip Normalization (AdaCN). As shown in \Cref{fig:short-b}, specifically, for a clip \( x \) with prompt frames, we first compute the mean \( \mu_y \) and standard deviation \( \sigma_y \) of the prompt frame features \( y \), and the mean \( \mu_x \) and standard deviation \( \sigma_x \) of the denoised prompt frames \( x_0 \). We then normalize the entire clip \( x \) using these statistics:

\begin{equation}
    \hat{x} = \sigma_y \left( \frac{x - \mu_x}{\sigma_x} \right) + \mu_y.
\end{equation}

Finally, we decode the concatenated feature sequences from multiple generated clips, after removing the redundant parts, into a coherent video.

\section{Experiments}

\subsection{Datasets}
\label{sec:datasets}
\noindent \textbf{Image Datasets.} We utilized two publicly available image-based try-on datasets—VITON-HD \cite{choi2021vitonhd} and DressCode \cite{morelli2022dresscode}—comprising 11,647 and 48,392 paired training samples, respectively, to construct our image-based training dataset. Comparative experiments for image-based virtual try-on were conducted on the test sets of VITON-HD \cite{choi2021vitonhd} and DressCode \cite{morelli2022dresscode} datasets.

\vspace{1mm}
\noindent \textbf{Video Datasets.} We constructed the ViViD-S dataset based on the ViViD \cite{fang2024vivid} dataset. Specifically, we trained a human orientation classifier based on EfficientNet \cite{tan2020efficientnetrethinkingmodelscaling} to detect sequences of frontal frames from the videos, filtering for videos that contain more than 24 consecutive frontal frames. Consequently, we selected 6,064 videos with a total of 513,896 frontal frames from the 7,759 videos in the ViViD \cite{fang2024vivid} training dataset to serve as our training set. Due to the impracticality of testing on thousands of videos from a time and cost perspective, we randomly selected 180 videos (60 for dresses, 60 for uppers, and 60 for bottoms) from the ViViD \cite{fang2024vivid} test set, each containing 64 consecutive frontal frames, to form the test set.  Additionally, we utilized the VVT \cite{fwgan-vvt} dataset, which is a standard video virtual try-on dataset comprising 791 paired person videos and clothing images, with a resolution of 192×256.
The comparative experiments are also conducted on the VVT test set which comprises 130 videos.

\subsection{Implementation Details} 
\label{imple_details} 

We initialized the DiT backbone using the pre-trained weights from EasyAnimateV4 \cite{xu2024easyanimatehighperformancelongvideo}, which was fine-tuned based on HunyuanDiT \cite{li2024hunyuandit}, and employed the pre-trained MagViT \cite{yu2023magvit} as the video VAE to handle video data.
We employed a progressive training strategy across three stages. In the first stage, we trained the model at a resolution of 256×192 using four datasets—DressCode, VITON-HD, VIViD, and VVT. Each training video sample consisted of 72 frames, with a batch size of 16, and the training ran for 128K steps. In the second stage, we increased the resolution to 512×384 and focused on three datasets—DressCode, VITON-HD, and VIViD. The number of frames per training sample was reduced to 48, with a batch size of 8, and the training continued for 64K steps. Finally, in the third stage, we used a high resolution of 832×624, training on the DressCode, VITON-HD, and VIViD-S datasets. In this stage, each sample contained 32 frames, the batch size was reduced to 1. This progressive strategy allowed the model to first learn at lower resolutions before refining its performance at higher resolutions.
For all stages, we applied the AdamW optimizer \cite{loshchilov2019adamw} with a constant learning rate of 1e-5 and gradient clipping set to 1.0. The entire training process was carried out on 4 NVIDIA A100 GPUs. Additionally, all model variants used in the ablation study were trained under identical hyperparameter settings to ensure fair comparison.

\subsection{Metrics}
For image-based try-on settings, we employ four widely used metrics: SSIM~\cite{wang2004ssim}, LPIPS~\cite{zhang2018LPIPS}, FID~\cite{Seitzer2020FID}, and KID~\cite{bińkowski2021kid}. SSIM and LPIPS are used to measure the similarity between two images, while FID and KID assess the similarity between two sets of images. In paired try-on tests, we use all four metrics, whereas in unpaired scenarios, we only use FID and KID.
For video try-on scenarios, we use SSIM ~\cite{wang2004ssim}, LPIPS~\cite{zhang2018LPIPS}, and VFID with I3D \cite{carreira2018i3d} and ResNext as metrics to evaluate video quality.
\subsection{Qualitative Comparison}


\begin{figure}
  \centering
  \includegraphics[width=0.48\textwidth]{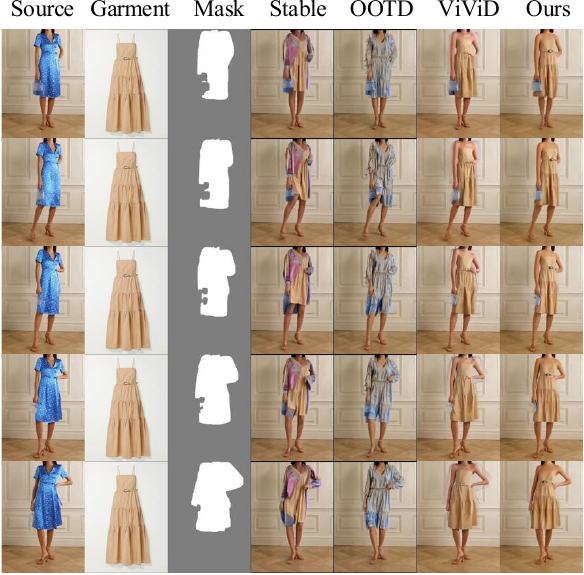}
  \caption{Qualitative comparison on the ViViD \cite{fang2024vivid} dataset for dresses. We use Stable and OOTD as the short for StableVITON \cite{kim2023stableviton} and OOTDiffusion \cite{xu2024ootdiffusion}. Additional comparison results are provided in the supplementary materials. Please zoom in for more details.}
  \label{fig:comparison-1}
\end{figure}

\begin{figure}
  \centering
  \includegraphics[width=0.48\textwidth]{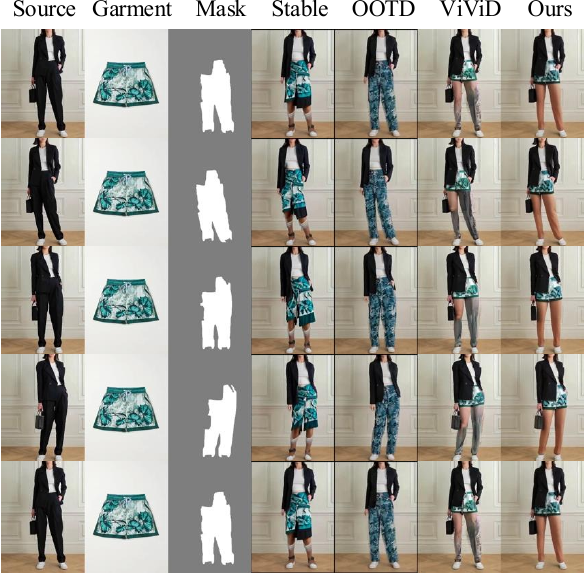}
  \caption{Qualitative comparison on the ViViD \cite{fang2024vivid} dataset for lower. We use Stable and OOTD as the short for StableVITON \cite{kim2023stableviton} and OOTDiffusion \cite{xu2024ootdiffusion}. Additional comparison results are provided in the supplementary materials. Please zoom in for more details.}
  \label{fig:comparison-2}
\end{figure}

\begin{figure}
  \centering
  \includegraphics[width=0.48\textwidth]{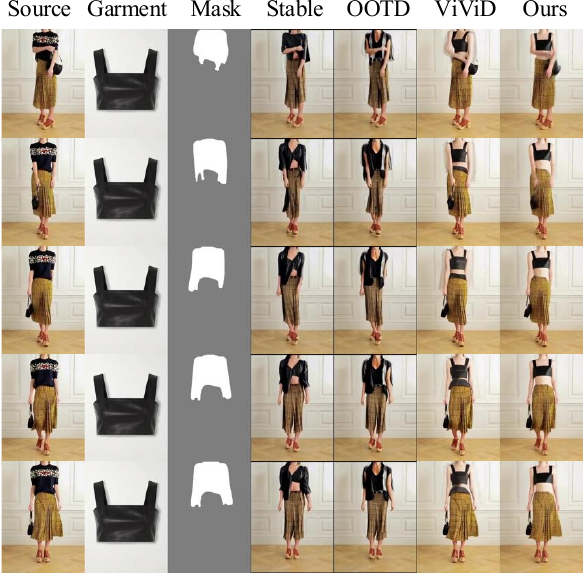}
  \caption{Qualitative comparison on the ViViD \cite{fang2024vivid} dataset for upper. We use Stable and OOTD as the short for StableVITON \cite{kim2023stableviton} and OOTDiffusion \cite{xu2024ootdiffusion}. Additional comparison results are provided in the supplementary materials. Please zoom in for more details.}
  \label{fig:comparison-3}
\end{figure}

\begin{figure*}
  \centering
  \includegraphics[width=\textwidth]{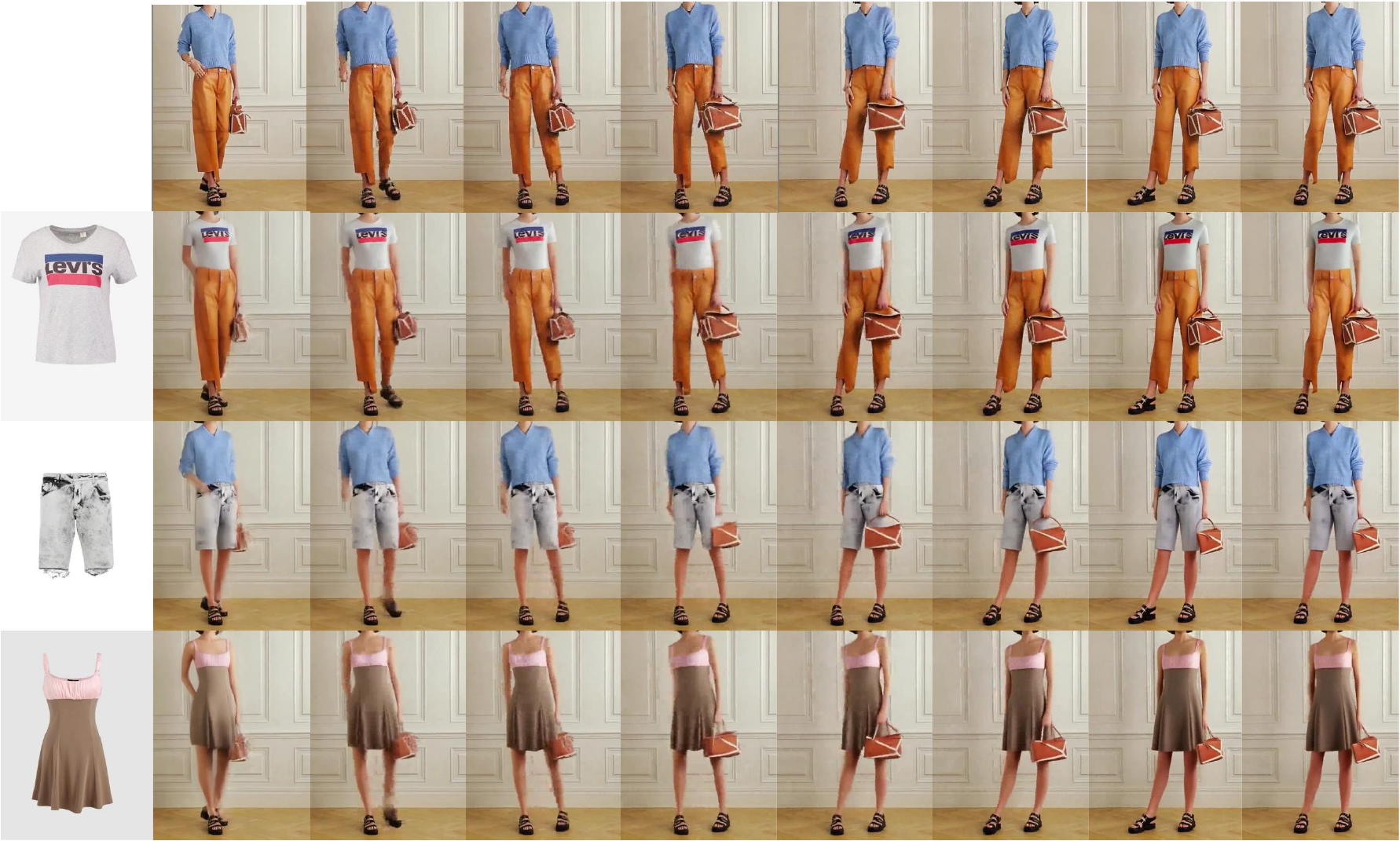}
  \caption{Results of video try-on with CatV$^2$TON. CatV$^2$TON can perform video try-on with various types of garments, achieving high consistency in garment texture and shape. Additional comparison results are provided in the supplementary materials. Please zoom in for more details.}
  \label{fig:result}
\end{figure*}

\Cref{fig:comparison-1}, \Cref{fig:comparison-2} and \Cref{fig:comparison-3} present qualitative comparisons of our method with StableVITON \cite{kim2023stableviton}, OOTDiffusion \cite{xu2024ootdiffusion}, and ViViD \cite{fang2024vivid} on the ViViD-S test set for unpaired visual try-on with dress, lower and upper clothing. Our approach demonstrates superior performance in generating try-on videos with improved temporal coherence and garment consistency compared to other baseline methods. 
\Cref{fig:result} presents the results of our method applied to the same person in a video with different types of clothing changes. This demonstrates the capability of our approach to maintain texture consistency, clothing shape, and temporal coherence.
Additional comparison results are provided in the supplementary materials.

\subsection{Quantitative Comparison}
\label{sec:quantity}

\noindent \textbf{Image-based Virtual Try-On.} We conducted quantitative comparisons with advanced image-based try-on methods \cite{gou2023dcivton, kim2023stableviton, wang2024stablegarment, wang2024mv, xie2023gpvton, morelli2023ladi-vton, xu2024ootdiffusion} under both paired and unpaired settings on the VITON-HD \cite{choi2021vitonhd} and DressCode \cite{morelli2022dresscode} datasets. 
During inference for image try-on, we used the DDPM \cite{ho2020ddpm} noise schedule for 20 steps, with a CFG \cite{ho2022cfg} strength set to 3.0.
As shown in \Cref{tab:vitonhd} and \Cref{tab:dresscode}, although our approach is designed for unified visual try-on, it outperforms traditional image-based try-on methods across various metrics for generated image quality, particularly in the unpaired scenario. This demonstrates that our method exhibits strong generalization performance even when trained on a single dataset.

\vspace{1mm}
\noindent \textbf{Video-based Virtual Try-On.}
Due to the limited availability of open-source video try-on methods, we selected two relatively high-performing image-based virtual try-on methods \cite{kim2023stableviton, xu2024ootdiffusion} for comparison. These methods were evaluated using frame-by-frame inference and by applying MagicAnimate \cite{xu2024magicanimate} to generate videos from the first frame. Additionally, we compared our approach with the video-based method ViViD \cite{fang2024vivid}. The comparison results on the ViViD-S test set are presented in \Cref{tab:vivid}, where our method outperforms others in both paired and unpaired scenarios across all evaluation metrics. 
We also compared our approach with other video try-on methods on the VVT \cite{fwgan-vvt} dataset, as shown in Table \Cref{tab:vvt}. Due to the limited availability of open-source implementations, we reproduced the results of ViViD \cite{fang2024vivid} using its code, while for other methods, we directly adopted the results reported in their papers. The results demonstrate that our method outperforms the others.
During inference for video try-on, we used the DDPM \cite{ho2020ddpm} noise schedule for 15 steps, with a CFG \cite{ho2022cfg} strength set to 2.5.

\begin{table}[t]
    \centering
    \resizebox{0.48\textwidth}{!}{
    \setlength{\tabcolsep}{1mm}
    \begin{tabular}{l|cccc|cc}
        \toprule
        \multirow{2}{*}{Methods} & \multicolumn{4}{c|}{Paired} & \multicolumn{2}{c}{Unpaired} \\
        \cline{2-7}
        & SSIM\ $\uparrow$  & FID $\downarrow$ & KID $\downarrow$ & LPIPS $\downarrow$  & FID $\downarrow$ & KID $\downarrow$ \\
        \midrule
        
        StableGarment~\cite{wang2024stablegarment} & 0.8029 & 15.567 & 8.519 & 0.1042 & 17.115 & 8.851  \\
        MV-VTON~\cite{wang2024mv} & 0.8083 & 15.442 & 7.501 & 0.1171 & 17.900 & 3.861 \\
        
        LaDI-VTON~\cite{morelli2023ladi-vton} & 0.8603 & 11.386 & 7.248 & 0.0733 & 14.648 & 8.754 \\
        DCI-VTON~\cite{gou2023dcivton} & 0.8620 & 9.408 & 4.547 & 0.0606 & 12.531 & 5.251 \\

        OOTDiffusion~\cite{xu2024ootdiffusion} & 0.8187 & 9.305 & 4.086 & 0.0876 & 12.408 & 4.689   \\
        GP-VTON~\cite{xie2023gpvton} & \underline{0.8701} & 8.726 & 3.944 & \underline{0.0585} & 11.844 & 4.310  \\
        StableVTON~\cite{kim2023stableviton} & 0.8543 & \textbf{6.439} & \textbf{0.942} & 0.0905 & \textbf{11.054} & \underline{3.914}   \\
        
        \rowcolor{LightBlue} CatV$^{2}$TON (Ours) & \textbf{0.8902} & \underline{8.095} & \underline{2.245} & \textbf{0.0570} & \underline{11.222} & \textbf{2.986} \\

        \bottomrule
    \end{tabular}
    }
    \caption{Quantitative comparison with other methods on VITON-HD \cite{choi2021vitonhd} dataset. The best and second-best results are demonstrated in \textbf{bold} and \underline{underlined}, respectively.}
    \label{tab:vitonhd}
\end{table}

\begin{table}[t]
    \centering
    \resizebox{0.48\textwidth}{!}{
    \setlength{\tabcolsep}{1mm}
    \begin{tabular}{l|cccc|cc}
        \toprule
        \multirow{2}{*}{Methods} & \multicolumn{4}{c|}{Paired} & \multicolumn{2}{c}{Unpaired}  \\
        \cline{2-7}
        & SSIM\ $\uparrow$  & FID $\downarrow$ & KID $\downarrow$ & LPIPS $\downarrow$  & FID $\downarrow$ & KID $\downarrow$  \\
        \midrule
        GP-VTON~\cite{xie2023gpvton} & 0.7711 & 9.927 & 4.610 & 0.1801 & 12.791 & 6.627 \\
        LaDI-VTON~\cite{morelli2023ladi-vton} & 0.7656 & 9.555 & 4.683 & 0.2366 & 10.676 & 5.787 \\
        IDM-VTON~\cite{choi2024idmvton} & 0.8797 & 6.821 & 2.924 & 0.0563 & \underline{9.546} & \underline{4.320}  \\
        OOTDiffusion~\cite{xu2024ootdiffusion} & \underline{0.8854} & \textbf{4.610} & \textbf{0.955} & \underline{0.0533} & 12.567 & 6.627 \\
        
        \rowcolor{LightBlue} CatV$^{2}$TON (Ours) & \textbf{0.9222} & \underline{5.722} & \underline{2.338} & \textbf{0.0367} & \textbf{8.627} & \textbf{3.838} \\

        \bottomrule
    \end{tabular}
    }
    \caption{Quantitative comparison with other methods on DressCode \cite{morelli2022dresscode} dataset. The best and second-best results are demonstrated in \textbf{bold} and \underline{underlined}, respectively.}
    \label{tab:dresscode}
\end{table}

\begin{table}[h]
    \centering
    \resizebox{0.48\textwidth}{!}{
    \setlength{\tabcolsep}{.7mm}
        \fontsize{9}{10} \selectfont
        \begin{tabular}{l|cccc|cc}
            \hline
            \multirow{2}{*}{Methods} & \multicolumn{4}{c|}{Paired} & \multicolumn{2}{c}{Unpaired}  \\
            \cline{2-7}
            &VFID$_{I}$ $\downarrow$ & VFID$_{R}$ $\downarrow$ &  SSIM $\uparrow$  &  LPIPS $\downarrow$ & VFID$_{I}$ $\downarrow$ & VFID$_{R}$ $\downarrow$  \\ 
            \midrule
            FW-GAN \cite{fwgan-vvt}     & 8.019  & 0.1215  & 0.675  & 0.283  & -      & -      \\ 
            MV-TON \cite{choi2024idmvton}   & 8.367  & 0.0972 & 0.853  & 0.233  & -      & -      \\ 
            ClothFormer \cite{jiang2022clothformer}  & 3.967  & 0.0505 & \textbf{0.921} & \underline{0.081}  & -      & -      \\
            WildVidFit \cite{he2024wildvidfit}   & 4.202  & -       & -      & -      & -      & -      \\ 
            ViViD  \cite{fang2024vivid}    & \underline{3.793}  & \underline{0.0348} & 0.822  & 0.107  & \underline{3.994}  & \underline{0.0416} \\
            \rowcolor{LightBlue} CatV$^{2}$TON (Ours)  & \textbf{1.778}  & \textbf{0.0103}  & \underline{0.900}  & \textbf{0.0385} & \textbf{1.902}  & \textbf{0.0141} \\ 
            \hline
        \end{tabular}
    }
    \caption{Quantitative comparison with other methods on VVT \cite{fwgan-vvt} dataset. The best and second-best results are demonstrated in \textbf{bold} and \underline{underlined}, respectively.}
    \label{tab:vvt}
\end{table}



\begin{table}[htbp]
    \centering
    \resizebox{0.48\textwidth}{!}{
    \setlength{\tabcolsep}{.7mm}
    \fontsize{9}{10} \selectfont
    \begin{tabular}{l|cccc|cc}
        \hline
        \multirow{2}{*}{Methods} & \multicolumn{4}{c|}{Paired} & \multicolumn{2}{c}{Unpaired}  \\
        \cline{2-7}
        & VFID$_{I}$ $\downarrow$ & VFID$_{R}$ $\downarrow$ &  SSIM $\uparrow$  &  LPIPS $\downarrow$ & VFID$_{I}$ $\downarrow$ & VFID$_{R}$ $\downarrow$  \\ 
        \midrule
        StableVITON \cite{kim2023stableviton}      & 34.2446 & 0.7735 & 0.8019  & 0.1338  & 36.8985 & 0.9064 \\ 
        OOTDiffusion \cite{xu2024ootdiffusion}                    & 29.5253 & 3.9372 & 0.8087  & 0.1232  & 35.3170 & 5.7078 \\ 
        IDM-VTON \cite{choi2024idmvton}                      & 20.0812 & \underline{0.3674} & 0.8227  & \underline{0.1163}  & 25.4972 & 0.7167 \\ 
        \midrule
        StableVITON+AM & 19.9239 & 0.7586 & 0.8207  & 0.1291  & 22.0262 & 0.8283 \\ 
        OOTDiffusion+AM & 19.3173 & 0.9382 & 0.8154  & 0.1244  & 23.3938 & 1.1485 \\ 
        IDM-VTON+AM & 18.2048 & 0.4481 & \underline{0.8252}  & 0.1212  & 22.5881 & \underline{0.5397} \\ 
        \midrule
        ViViD \cite{fang2024vivid}          & \underline{17.2924} & 0.6209 & 0.8029  & 0.1221  & \underline{21.8032} & 0.8212 \\ 
        \rowcolor{LightBlue} CatV$^{2}$TON (Ours) & \textbf{13.5962} & \textbf{0.2963} & \textbf{0.8727}  & \textbf{0.0639}  & \textbf{19.5131} & \textbf{0.5283} \\ \hline
    \end{tabular}
    }
    \caption{Quantitative comparison with other methods on ViViD dataset. The best and second-best results are demonstrated in \textbf{bold} and \underline{underlined}, respectively.}
    \label{tab:vivid}
\end{table}

\subsection{Ablation Studies}
\label{sec:ablation}

\begin{figure}
  \centering
  \includegraphics[width=0.48\textwidth]{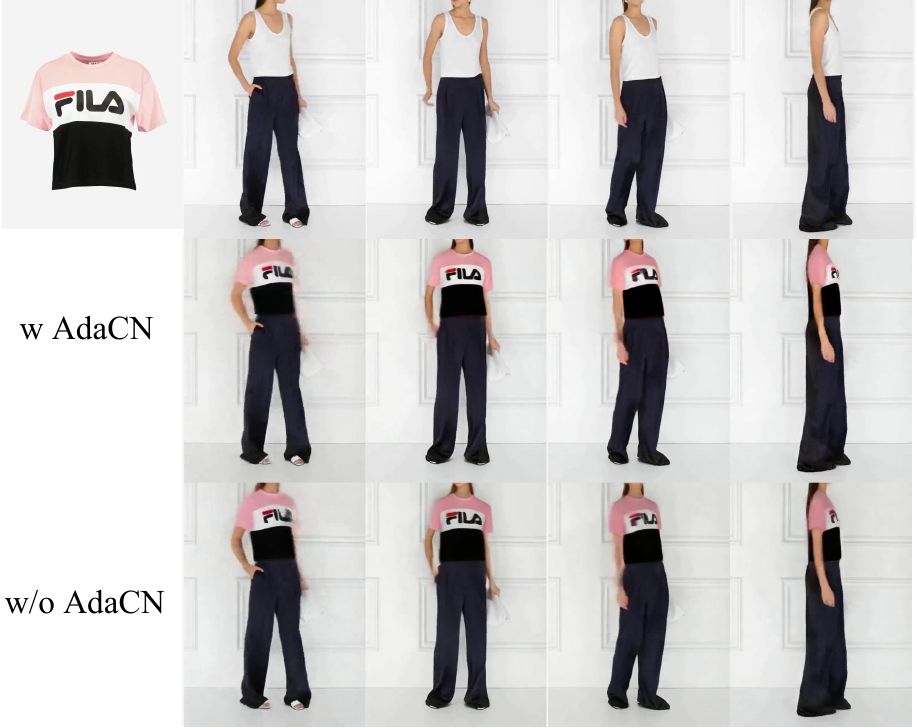}
  \caption{Ablation visual results about AdaCN. When AdaCN is not utilized for inference, the clothing parts in the try-on results will exhibit color difference issues, which typically intensify with the increase in video length.}
  \label{fig:ablation}
\end{figure}

To validate the contribution of different components or strategies to the final performance, we conducted ablation experiments on PoseNet, training data (ViViD \cite{fang2024vivid} dataset or ViViD-S dataset), and AdaCN.

\vspace{1mm}
\noindent \textbf{PoseEncoder.} We trained a model without PoseEncoder, using the same hyperparameters as mentioned in \Cref{imple_details}, to assess the impact of adding PoseEncoder on performance. As shown in \Cref{tab:trainable_module} and \Cref{fig:ablation}, the performance decreased without pose information, and it exhibited defects in handling complex poses.


\vspace{1mm}
\noindent \textbf{AdaCN.} We introduced Adaptive Clip Normalization (AdaCN) for long video stitching inference in \Cref{subsec:infer}. To validate its effectiveness, we conducted ablation experiments to assess alignment from both metric and visual perspectives. As shown in \Cref{fig:ablation}, without AdaCN, the inference results showed noticeable color mismatch accumulation, and the transitions between segments were not smooth. \Cref{tab:trainable_module} further indicates that the use of AdaCN improves the quality of generated videos.

        
        
\begin{table}[t]
    \centering
    \resizebox{0.48\textwidth}{!}{
    \setlength{\tabcolsep}{1.mm}
    \fontsize{9}{10} \selectfont
        \begin{tabular}{c|ccccc|cc}
        \toprule
        \multirow{2}{*}{Variations} & \multicolumn{5}{c|}{Paired} & \multicolumn{2}{c}{Unpaired} \\
        \cline{2-8}
        & VFID$_{I}$ $\downarrow$ & VFID$_{R}$ $\downarrow$ & SSIM $\uparrow$ & LPIPS $\downarrow$ &  & VFID$_{I}$ $\downarrow$ & VFID$_{R}$ $\downarrow$  \\ 
        \midrule
        w/o Pose    & 4.6398  & 0.0562  & 0.8280  & 0.0990  & & 4.9894 & 0.0598 \\
        w/o AdaCN   & \underline{4.5125}  & \textbf{0.0438}  & \textbf{0.8507}  & \textbf{0.0842}  & & \underline{4.4474} & \underline{0.0522} \\
        \rowcolor{LightBlue} CatV$^{2}$TON & \textbf{4.3657} & \underline{0.0491} & \underline{0.8461} & \underline{0.0983} & & \textbf{4.4324} & \textbf{0.0506} \\	 	 	 
        \bottomrule
        \end{tabular}
    }
    \caption{Ablation results of AdaCN, PoseEncoder. The best and second-best results are demonstrated in \textbf{bold} and \underline{underlined}, respectively.}
    \label{tab:trainable_module}
\end{table}


\section{Limitations}
\noindent \textbf{Resolution and Clarity.}  
In comparison to image data, video data, despite having a similar resolution, cannot achieve the same level of clarity due to the dynamic changes it exhibits. Even though the current resolution of 832$\times$624 is sufficient in terms of pixel count, it still falls short of meeting application requirements in terms of clarity. A higher-quality, higher-resolution video try-on dataset is crucial to addressing this issue.

\vspace{1mm}
\noindent \textbf{Physical Laws in Video.}  
The key difference between video and image data lies in the dynamic nature of video, which must adhere strictly to physical laws. Otherwise, unrealistic artifacts may emerge, particularly in try-on tasks where clothing movement during different actions is critical. Currently, there is a lack of foundational video generation models that can accurately simulate these physical behaviors. Achieving this would likely require larger-scale models and a higher-quality, larger-volume video dataset.
\section{Conclusion}
\label{sec:conclusion}

In this work, we proposed CatV$^2$TON, a simple and efficient diffusion transformer framework for both image and video virtual try-on tasks. By temporally concatenating garment and person inputs and training with a mixed image-video dataset, our model achieves high-quality results with only 20\% of the backbone parameters as trainable components. To support long, temporally consistent try-on video generation, we introduced an overlapping clip-based inference strategy with Adaptive Clip Normalization (AdaCN), reducing resource demands while maintaining temporal continuity. Additionally, we propose a curated video try-on dataset, ViViD-S, created by filtering out back-view frames and applying 3D Mask Smoothing to enhance the temporal consistency of masks. Extensive experiments demonstrate that CatV$^2$TON outperforms baseline methods in both quantitative and qualitative evaluations, marking a significant step forward for unified models in vision-based virtual try-on research.

{
    \small
    \bibliographystyle{ieeenat_fullname}
    \bibliography{main}
}


\end{document}